\documentclass[twoside,11pt]{article}
\usepackage{enumitem}
\usepackage{blindtext}
\usepackage{amsmath}
\usepackage{bbm}
\usepackage[T1]{fontenc}
\usepackage{subcaption}

\usepackage[preprint]{jmlr2e}

\usepackage{xcolor}
\usepackage{tabularx}

\usepackage{makecell}
\usepackage{booktabs}
\usepackage{lastpage}

\firstpageno{1}

\begin{document}

\title{RouteJudge: An Open Platform for Reproducible and Preference-Aware LLM Routing}

\author{\name Guannan Lai \textsuperscript{1,2,3} \email laign@lamda.nju.edu.cn 
       \\
       \name Haoran Hu \textsuperscript{1,2} \email huhr@smail.nju.edu.cn  
       \\ 
       \name Han-Jia Ye \textsuperscript{1,2} \email yehj@lamda.nju.edu.cn \\
       \addr \textsuperscript{1}School of Artificial Intelligence, Nanjing University, China\\
       \textsuperscript{2}National Key Laboratory for Novel Software Technology, Nanjing University, 210023, China\\
       \textsuperscript{3}SinapisAI
}
\editor{My editor}

\maketitle

\begin{abstract}
LLM routing aims to automatically select the most suitable model from a heterogeneous model pool for each incoming query, so as to optimize routing quality under practical constraints such as cost, latency, and throughput. However, existing evaluations of large language model (LLM) routing systems typically rely on static benchmarks, golden answers, or automated quality scores, which impose a fixed notion of response quality and overlook the pluralistic user preferences that arise in real-world interactions. 

We present \textsc{\textbf{RouteJudge}}, an online pairwise preference evaluation framework for LLM routing systems, with a public platform available at \url{https://routejudge.cn}. Different from model-level response evaluation, \textsc{\textbf{RouteJudge}} focuses on router-level decision quality. For each user query, multiple routing strategies independently recommend candidate models under the same model pool and budget constraints. The selected model responses are then presented to users through anonymous pairwise comparisons, and the resulting user preferences are attributed back to the routing strategies behind the compared responses. Each evaluation record further stores the query, routing decisions, model responses, preference labels, cost, latency, and task metadata, enabling preference-aware, cost-aware, and task-conditioned analysis of LLM routers.

To support the continuous expansion of routing methods in \textsc{RouteJudge}, we further release \textsc{ORBIT} (Optimal Routing and Budgeted Inference Toolbox), a modular and extensible toolbox that standardizes the end-to-end workflow of LLM routing. \textsc{ORBIT} provides unified interfaces for benchmark loading, query representation, router implementation, budget-aware evaluation, and method comparison, allowing researchers to develop and evaluate routing algorithms under consistent protocols. More importantly, \textsc{ORBIT} serves as the submission and integration layer for \textsc{RouteJudge}: researchers can implement their routing methods within \textsc{ORBIT}, evaluate them on existing routing benchmarks, and submit compatible routers for online preference-based evaluation on \textsc{RouteJudge}. The code of \textsc{ORBIT} is available at \url{https://github.com/LAMDA-Model-Reuse/ORBIT}.
\end{abstract}

\begin{keywords}
  Large Language Model, Routing, Budget, Deep Learning, Efficient Inference
\end{keywords}

\section{Introduction}
Large language models (LLMs) are increasingly deployed not as a single monolithic system, but as a heterogeneous pool of proprietary and open-source models with distinct capabilities, costs, latency profiles, context lengths, modalities, and serving characteristics \citep{wang2025mixllm,aggarwal2024automix}. In real-world applications, the optimal model is often query-dependent: simple requests may be answered by cheaper or faster models, while complex, high-stakes, or preference-sensitive queries may require stronger models to achieve satisfactory utility. This has motivated \emph{LLM routing}, which aims to automatically select the most suitable model for each incoming query under practical constraints such as cost, latency, and throughput \citep{feng2025graphrouter,chen2023frugalgpt,varangot2025doing}. Beyond maximizing raw accuracy, routing must reason about budget-induced trade-offs, service-level objectives, uncertainty, and failure behaviors such as cascading fallback, making it a central mechanism for scalable and cost-effective LLM deployment \citep{srivatsa2024harnessing,wang2025icl}. From a broader perspective, LLM routing also resonates with the learnware vision of model reuse in a model market: instead of training a bespoke model for every scenario, users can reuse and compose suitable pre-trained models according to query-side information and deployment constraints \citep{zhou2016learnware,zhou2024learnware,tan2024beimingwu}. We therefore view routing as a form of \emph{budgeted inference}, where a router must select a budget-feasible model for each query while maximizing expected user-facing utility. This setting naturally induces a performance--cost trade-off curve and calls for evaluation protocols that go beyond a single static operating point.

Despite the growing importance of LLM routing, its evaluation remains largely dominated by offline protocols. Existing methods are typically assessed using static benchmarks, golden answers, automated scores, or routing accuracy against an assumed best model~\citep{hu2024routerbench,huang2025routereval}. While these protocols are convenient and reproducible, they reduce routing quality to a fixed objective target and therefore provide only a partial view of deployment-time performance. In realistic interactions, many queries do not admit a single universally optimal response: for creative writing, translation, tutoring, analytical reasoning, and dialogue, multiple outputs may be valid, yet users may prefer different responses depending on their expectations, cost sensitivity, latency tolerance, desired level of detail, reasoning style, or tone~\citep{kirk2024prism,sorensen2024value,feng2024modular}. As a result, a router that performs well under golden-answer or automated evaluation may not select models whose responses are actually preferred by users. We view this mismatch as a pluralistic preference alignment problem for LLM routing, where routers should be evaluated not only by whether they select a benchmark-optimal model, but also by whether their decisions lead to user-preferred responses under realistic query distributions.

To address this limitation, we propose \textbf{RouteJudge}, an online platform for evaluating LLM routers under pluralistic user preferences. Rather than comparing routing decisions against fixed golden answers or an assumed best model, RouteJudge evaluates whether a router selects models whose responses are actually preferred by users. For each query, multiple routing strategies operate under the same model pool and budget constraints, and the outputs of their selected models are judged through anonymous pairwise comparisons. The resulting preference signal is attributed back to the routers behind the compared responses, shifting the evaluation target from model-level response quality to router-level decision quality. By recording queries, routing decisions, preferences, costs, latencies, and task metadata, RouteJudge supports preference-aware, cost-aware, and task-conditioned analysis of routing behavior in realistic user-facing settings.

To continuously expand the set of routing methods evaluated by \textbf{RouteJudge}, we further present \textsc{ORBIT} (\textbf{O}ptimal \textbf{R}outing and \textbf{B}udgeted \textbf{I}nference \textbf{T}oolbox), a modular and extensible toolbox for standardizing the end-to-end workflow of LLM routing. \textsc{ORBIT} provides unified interfaces for benchmark loading, query representation, router implementation, budget-aware evaluation, and method comparison, enabling the development and assessment of different routing algorithms under consistent protocols. More importantly, ORBIT serves as the submission and integration layer for RouteJudge: researchers can implement their routing methods within \textsc{ORBIT}, evaluate them on existing routing benchmarks, and submit compatible routers for online preference-based evaluation on RouteJudge. In this way, \textsc{ORBIT} and RouteJudge together form an open evaluation ecosystem, where routing methods can first be developed and validated offline under reproducible settings, and then tested online under real user preferences.

\section{Background and Motivation}

\subsection{LLM Routing}

LLM routing aims to coordinate a pool of language models with different capabilities, costs, and response characteristics. Given a user query and its surrounding context, a router selects one or more candidate models to answer the query, with the goal of improving the trade-off among response quality, inference cost, latency, and task-specific requirements~\citep{shnitzer2023llm,lai2026routing}. Compared with always using a single fixed model, routing provides a more flexible inference paradigm: easier queries can be assigned to cheaper or faster models, while more difficult or preference-sensitive queries can be assigned to stronger models \citep{varangot2025doing,lai2026sampled}.

Existing routing methods instantiate this idea in different ways. Similarity-based routers estimate model suitability by comparing the current query with previously evaluated examples in an embedding space~\citep{zhuang2024embedllm,reimers2019sentence}. Learned cost-quality routers train predictive models to estimate the expected utility of each candidate model under a budget constraint~\citep{ding2024hybrid,ong2024routellm,vsakota2024fly}. Cascade and uncertainty-based routers first query a cheaper model and invoke a stronger model only when confidence is low or the initial response appears insufficient~\citep{jiang2023llm,aggarwal2024automix}. Preference-based and structured routers further incorporate human or model preference signals, graph relations, or ensemble-style decisions to improve model selection~\citep{ramirez2024optimising,yue2023large}.

Although these methods differ in routing mechanism, they usually share a common evaluation assumption: routing quality can be measured offline using benchmark labels, task-specific metrics, or automated judges~\citep{hu2024routerbench,huang2025routereval,ma2026mmrbench}. Under this protocol, a router is considered effective if the model it selects obtains a high score on a fixed benchmark. This assumption is convenient for reproducible comparison, but it also narrows the evaluation target. It measures whether a router selects the benchmark-optimal model, rather than whether the routing decision matches what real users would prefer under heterogeneous quality, cost, and latency expectations.

\subsection{Why Offline Routing Evaluation Is Insufficient}

Offline evaluation offers a controlled and reproducible way to compare routing methods, but it only partially reflects routing quality in user-facing deployment. Existing protocols usually measure whether a router agrees with fixed labels, static benchmark scores, automated judges, or an assumed best model. This reduces routing to a fixed-target prediction problem, whereas a router in practice makes a deployment decision: which model should answer a query under a specific task context, user objective, and resource constraint.

This abstraction becomes inadequate for open-ended and preference-sensitive tasks. In writing, dialogue, translation, summarization, tutoring, and analytical reasoning, multiple responses may be factually valid while differing in tone, structure, level of detail, reasoning style, and presentation. Automatic metrics and benchmark labels can capture part of response quality, but they cannot fully determine which response a user will prefer in a particular interaction~\citep{zhang2019bertscore,bevilacqua2025automated}. Moreover, preferences are heterogeneous: some users favor concise answers, others prefer detailed explanations; some are sensitive to cost or latency, while others prioritize reasoning depth, formatting, or stylistic fit~\citep{kirk2024prism,sorensen2024value,feng2024modular}. As a result, a router that performs well under offline correctness or automated scoring may still select models whose outputs are not preferred by users.

The limitation is therefore not only about evaluation data, but also about the evaluation objective. LLM routing is inherently a multi-objective decision problem over quality, cost, latency, and user preference, while offline benchmarks often report quality as the primary score and treat cost as an auxiliary statistic~\citep{chen2023frugalgpt,ding2024hybrid}. A useful routing decision should produce responses that users prefer under realistic query distributions and deployment constraints, not merely match a benchmark-optimal model. \textbf{RouteJudge} is designed around this principle: it collects blinded pairwise preference feedback for model outputs selected by different routers and attributes the resulting preferences back to router-level decisions, shifting routing evaluation from fixed-target offline scoring to preference-aware assessment.

\begin{figure*}[t]
  \centering
  \includegraphics[width=\linewidth]{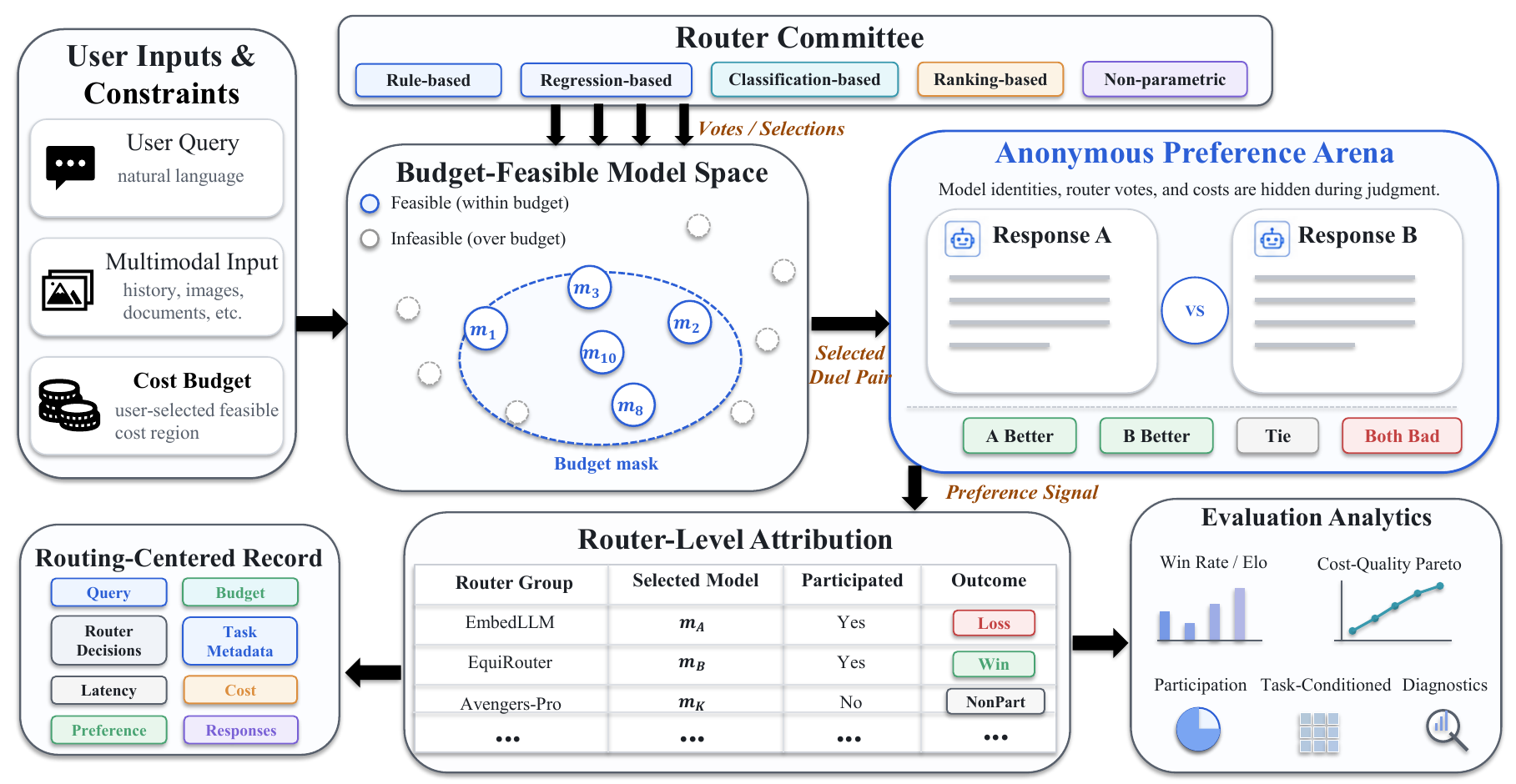}
  \caption{
  Overview of the RouteJudge evaluation framework. Given a user query, optional multimodal input, and a user-selected cost budget, a committee of routing strategies recommends models from the budget-feasible model space. RouteJudge selects a duel pair, presents the two responses through an anonymous preference interface, and attributes the resulting preference signal back to the routing strategies behind the compared model choices. Each interaction is stored as a routing-centered record containing the query, budget, router decisions, selected responses, preference label, cost, latency, and task metadata, supporting downstream analyses such as win rate, Elo rating, cost-quality Pareto analysis, participation, and task-conditioned diagnostics.
  }
  \label{fig:scoring}
\end{figure*}

\section{The RouteJudge Evaluation Framework}

\textbf{RouteJudge} is an online platform for evaluating LLM routing strategies through user preference feedback. Unlike conventional routing benchmarks that compare routers against fixed reference answers or offline model scores, RouteJudge evaluates whether routing decisions lead to responses that users actually prefer. Each router is treated as a black-box decision maker: given the same query, model pool, and budget constraint, it recommends one candidate model. RouteJudge then compares the responses produced by selected models through anonymous pairwise judgments and attributes the resulting preference signal back to the routers that made the corresponding decisions.

This section focuses on the evaluation protocol of RouteJudge. The training, implementation, offline benchmarking, and submission of routing methods are handled by \textsc{ORBIT}, which will be introduced in the next section as the standardized development and integration layer.

\subsection{Evaluation Record}

Each RouteJudge interaction is stored as a routing-centered decision record rather than as an isolated response-quality annotation. Formally, an evaluation record is written as
\[
\mathcal{Z}
=
\left(
x, h, I, C, \mathcal{M}_C,
\mathbf{d}, \mathbf{v},
m_A, m_B,
y,
\mathbf{s},
\mathbf{c},
\boldsymbol{\ell},
\tau,
\eta
\right),
\]
where $x$ is the user query, $h$ is the optional conversation history, $I$ is the optional multimodal input, $C$ is the user-selected budget, and $\mathcal{M}_C$ is the budget-feasible model set. The vector $\mathbf{d}$ records the models selected by all routers, and $\mathbf{v}$ denotes the induced model-level vote distribution. The pair $(m_A,m_B)$ is the model pair shown to the user, $y$ is the user preference label, and $\mathbf{s}$ stores the router-level attribution scores. The vectors $\mathbf{c}$ and $\boldsymbol{\ell}$ record inference costs and latencies, $\tau$ denotes the task label, and $\eta$ contains additional metadata.

This record structure makes RouteJudge different from model-level leaderboards. It allows the platform to analyze not only which response wins a comparison, but also which routers selected the winning model, how often each router participates in evaluated comparisons, how routing behavior changes under different budgets, and how performance varies across task types.

\subsection{Evaluation Workflow}

Figure~\ref{fig:scoring} summarizes the RouteJudge evaluation pipeline. For each user query, RouteJudge follows five main stages.

\textbf{Query and budget submission.}
The user submits a text-only or multimodal query and selects a cost budget $C$. The budget determines which candidate models are feasible for the current query. Given the full model pool $\mathcal{M}$, RouteJudge constructs the budget-feasible model set
\[
\mathcal{M}_C
=
\{m \in \mathcal{M} \mid \hat{c}(m \mid x,h,I) \leq C\},
\]
where $\hat{c}(m \mid x,h,I)$ denotes the estimated inference cost of model $m$ on the current input.

\textbf{Router recommendation.}
All routing strategies receive the same query context and the same feasible model set. Let $\mathcal{R}=\{r_1,r_2,\ldots,r_N\}$ denote the evaluated router set. Each router independently selects one model:
\[
m_i = r_i(x,h,I,\mathcal{M}_C),
\quad
m_i \in \mathcal{M}_C.
\]
RouteJudge does not assume any specific internal form of $r_i$; the router may be rule-based, retrieval-based, learned, preference-based, or otherwise implemented.

\textbf{Duel model selection.}
RouteJudge aggregates router decisions into model-level votes. For each model $m\in\mathcal{M}_C$, its vote count is
\[
v(m)=\left|\{i \mid m_i=m\}\right|.
\]
The platform selects two distinct models with the highest vote counts as the duel pair $(m_A,m_B)$. When multiple models are tied, RouteJudge prioritizes models with fewer historical comparisons to improve evaluation coverage, and breaks remaining ties randomly.

\textbf{Anonymous pairwise judgment.}
The two selected models generate responses to the same query in parallel. Their responses are presented to the user in randomized anonymous order, with model identities, router decisions, vote counts, costs, and latencies hidden during judgment. The user then provides one of four labels:
\[
y \in
\{
\text{\texttt{A Win}},
\text{\texttt{B Win}},
\text{\texttt{Tie}},
\text{\texttt{Both Bad}}
\}.
\]
This label space avoids forcing a binary preference when the two responses are indistinguishable or both unsatisfactory.

\textbf{Result reveal and router attribution.}
After the user submits the preference label, RouteJudge reveals the model identities, router vote distribution, costs, latencies, and router-level outcomes. The preference label is first converted into comparison scores:
\begin{equation}
(s_A,s_B)=
\begin{cases}
(1,0), & y=\text{\texttt{A Win}},\\
(0,1), & y=\text{\texttt{B Win}},\\
(0.5,0.5), & y=\text{\texttt{Tie}},\\
(0,0), & y=\text{\texttt{Both Bad}}.
\end{cases}
\label{eq:preference_score}
\end{equation}
The scores are then attributed to routers according to whether their selected models entered the evaluated duel:
\[
S_i =
\begin{cases}
s_A, & m_i=m_A,\\
s_B, & m_i=m_B,\\
\varnothing, & m_i\notin\{m_A,m_B\}.
\end{cases}
\]
Here, $\varnothing$ indicates that router $r_i$ is non-participating in this comparison and is not counted as either a win or a loss.

\begin{figure*}[t]
  \centering
  \begin{subfigure}[t]{0.49\linewidth}
    \centering
    \includegraphics[width=\linewidth]{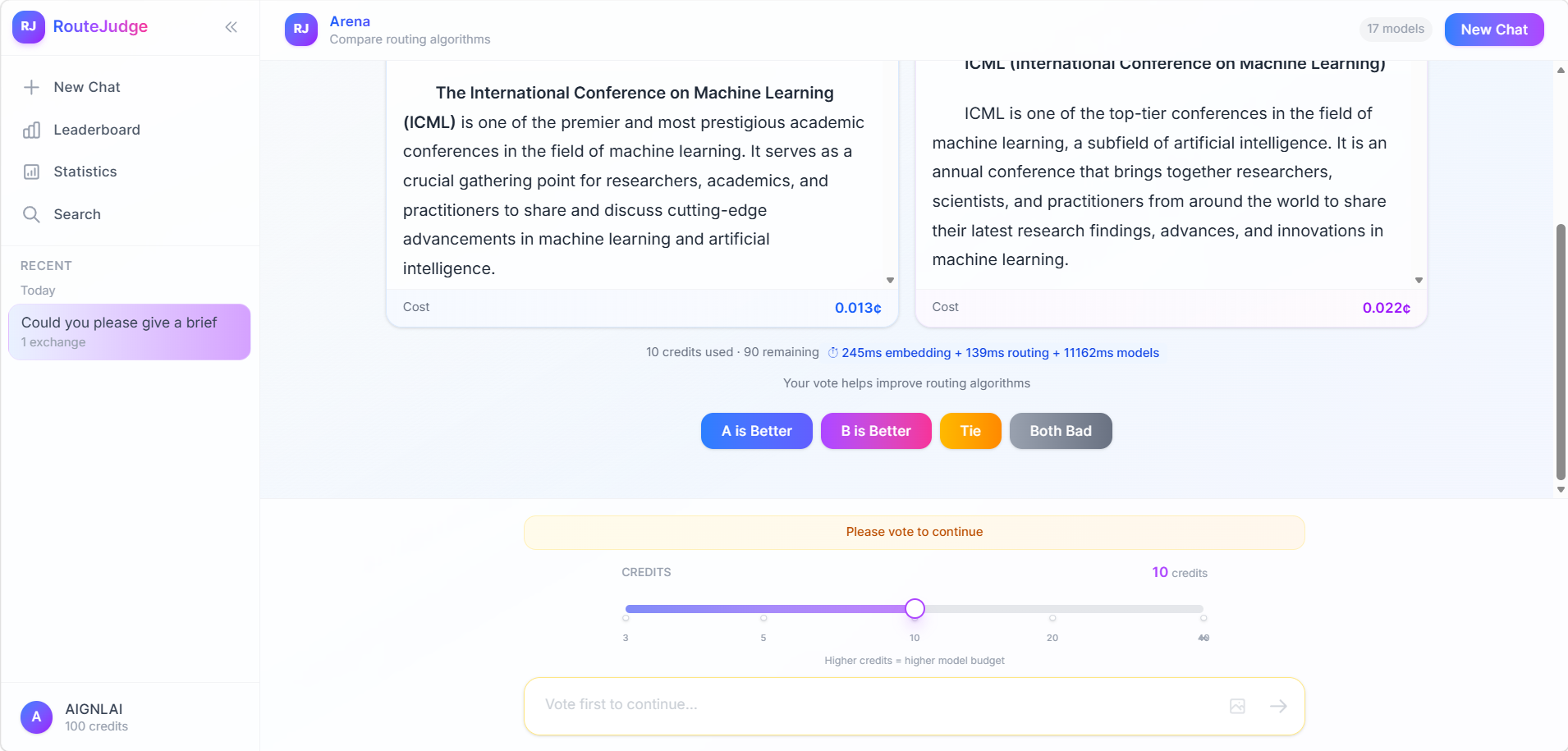}
    \caption{
    Anonymous pairwise preference interface in RouteJudge. Two model responses are presented side by side in randomized order, with model identities and router votes hidden during judgment. Users provide one of four labels: A Win, B Win, Tie, or Both Bad.
    }
    \label{fig:pipeline}
  \end{subfigure}
  \hfill
  \begin{subfigure}[t]{0.49\linewidth}
    \centering
    \includegraphics[width=\linewidth]{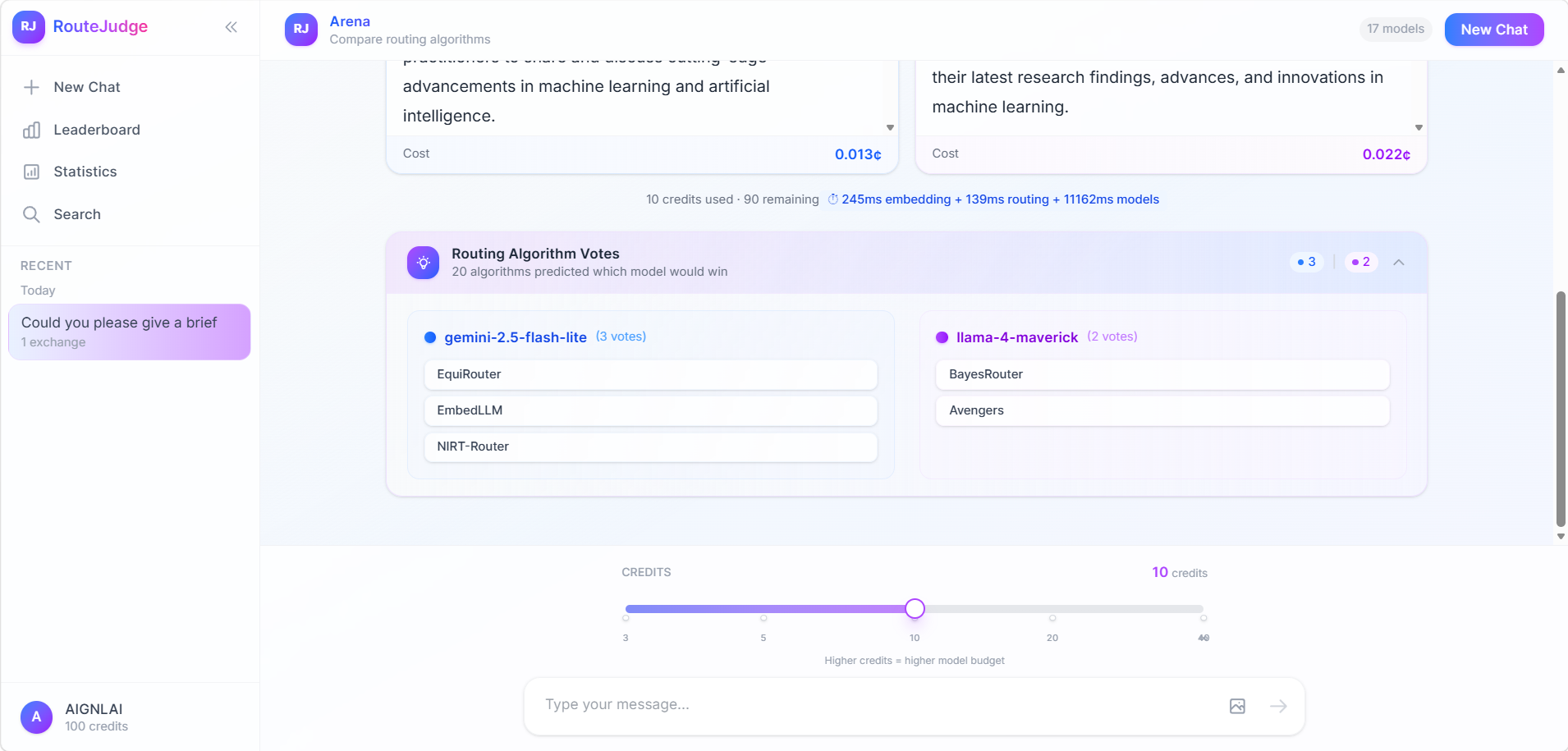}
    \caption{
    Result reveal and router attribution interface in RouteJudge. After the user submits a preference label, RouteJudge reveals the model identities, displays the routing vote distribution, reports cost and latency information, and assigns router-level outcomes according to whether each router selected the user-preferred model.
    }
    \label{fig:voted}
  \end{subfigure}
  \caption{
  RouteJudge user interface. Left: anonymous pairwise preference interface used for blinded user judgment. Right: result reveal and router attribution interface shown after preference submission.
  }
  \label{fig:routejudge_ui}
\end{figure*}

\subsection{Task Metadata and Evaluation Coverage}

To support task-conditioned analysis, RouteJudge assigns each query a task label $\tau$. In the current implementation, queries are mapped to broad categories such as Coding, Math, Translation, Creative Writing, Analysis, and Other. This metadata allows RouteJudge to report not only overall router performance, but also how routing behavior changes across heterogeneous task types.

RouteJudge also records whether each router participates in a judged comparison. A router may recommend a model that does not enter the selected duel pair; in this case, it receives $\varnothing$ rather than a positive or negative outcome. This design avoids assigning credit or blame to routers whose selected models were not actually judged by the user. At the same time, non-participation is itself informative: a router with high preference scores but low participation may have limited evaluation coverage, while a router with broader participation may provide a more reliable estimate of deployment-time behavior. RouteJudge therefore reports participation rate together with preference-based metrics.
\section{ORBIT for LLM Routing}

The RouteJudge evaluation protocol assumes that a diverse set of routers can be invoked under the same model pool, budget constraint, and prediction interface. However, continuously expanding this router pool is non-trivial: different routing methods often rely on different data formats, query encoders, training procedures, configuration files, and evaluation scripts. Without a standardized development layer, adding a new router to RouteJudge would require substantial engineering effort and could introduce uncontrolled differences in preprocessing, model access, or budget handling.

To make RouteJudge open and continuously extensible, we introduce \textsc{ORBIT} (\textbf{O}ptimal \textbf{R}outing and \textbf{B}udgeted \textbf{I}nference \textbf{T}oolbox), a modular toolbox that standardizes the end-to-end workflow of LLM routing. \textsc{ORBIT} provides unified interfaces for data loading, query representation, router implementation, training, validation, and deployment. In our ecosystem, \textsc{ORBIT} serves as the development and submission layer for RouteJudge: researchers can implement their routing algorithms within \textsc{ORBIT}, verify that they are directly runnable under a shared protocol, and submit compatible routers for historical replay and online preference-based evaluation on RouteJudge.

\begin{figure}[htb]
	\centering
	\includegraphics[width=\linewidth]{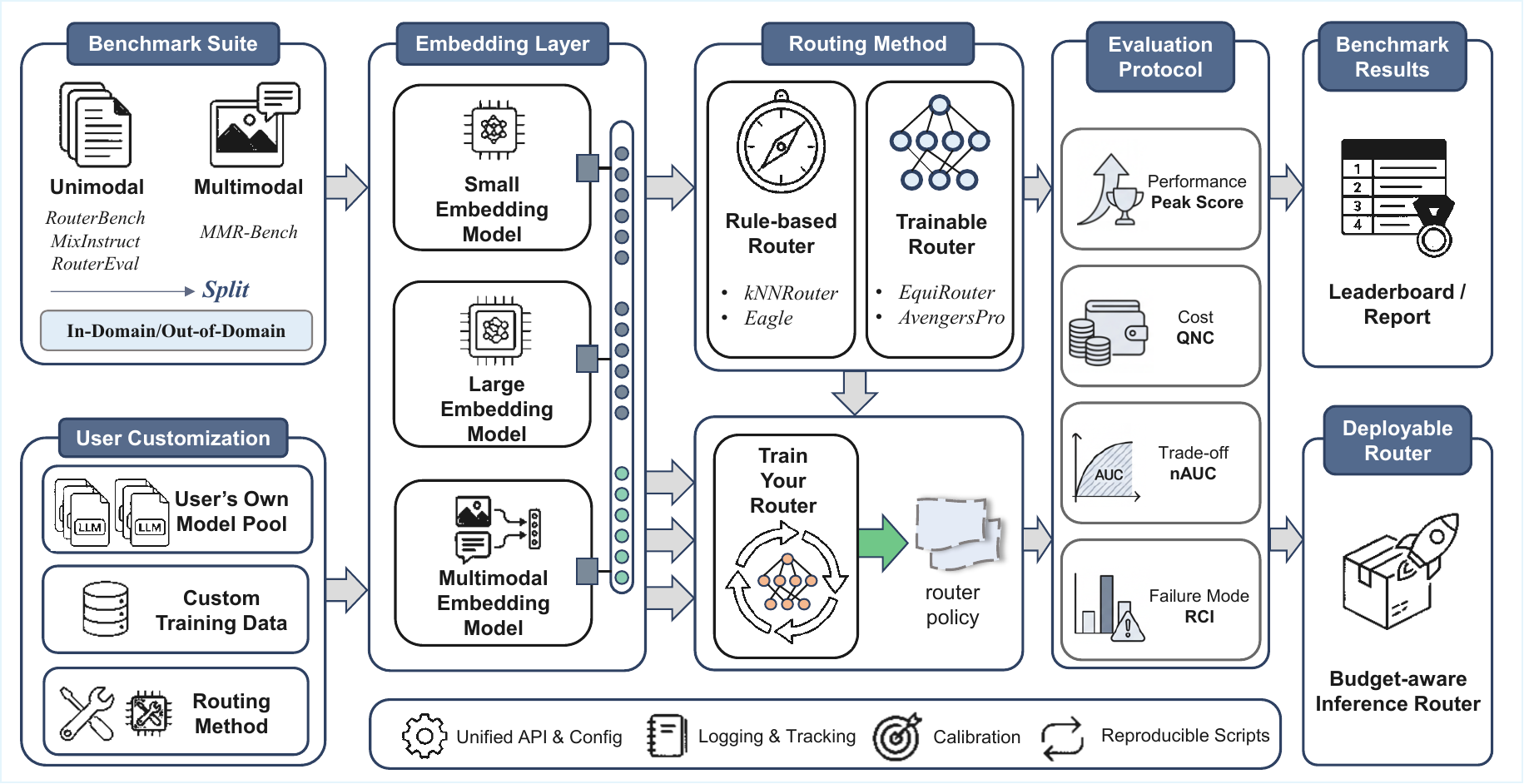}
	\caption{
	Overview of \textsc{ORBIT}, a modular end-to-end pipeline for LLM routing. 
	\textsc{ORBIT} unifies benchmark or user-provided data, query representation, router implementation, training, offline validation, and deployment, and further serves as the integration layer through which external routing methods can be submitted to \textbf{RouteJudge}.
	}
	\label{fig:orbit}
\end{figure}

\subsection{Overview of the ORBIT Pipeline}

\textsc{ORBIT} (\textbf{O}ptimal \textbf{R}outing and \textbf{B}udgeted \textbf{I}nference \textbf{T}oolbox) is the standardized development and integration layer behind \textbf{RouteJudge}. While RouteJudge focuses on online preference-based evaluation, \textsc{ORBIT} provides the infrastructure needed to implement, train, validate, and submit routing methods under a unified interface. As shown in Figure~\ref{fig:orbit}, the pipeline is organized around a set of composable components, including data sources, query representations, routing policies, training utilities, evaluation scripts, and deployment adapters.

The pipeline starts from either built-in routing benchmarks or user-provided data. Users may specify their own model pool, training data, routing method, and query encoder. \textsc{ORBIT} then converts each query into a shared representation through its embedding layer, which supports lightweight text encoders, larger embedding models, and multimodal encoders. This design separates query representation from router implementation, allowing different routers to be compared under the same input features or the same router to be tested with different encoders.

Given the query representation, a routing method is trained or initialized through a unified router interface. \textsc{ORBIT} supports different types of routers, including rule-based, retrieval-based, ranking-based, classification-based, regression-based, and trainable neural routers. Regardless of their internal design, all routers expose the same prediction interface: given a query and a feasible model set, the router outputs the model it recommends. This abstraction makes routers directly reusable across offline experiments and online RouteJudge evaluation.

After training, \textsc{ORBIT} runs standardized validation and logging. It records the dataset configuration, encoder choice, router hyperparameters, random seed, budget setting, model pool, and evaluation outputs. The goal is not only to compare routing methods, but also to make each result reproducible and each router deployable. Once a router satisfies the required interface and runtime constraints, it can be exported as a RouteJudge-compatible router and evaluated on the RouteJudge platform.

\subsection{Using ORBIT}

\textsc{ORBIT} is designed around two principles: composability and reproducibility. Composability means that datasets, encoders, routers, and evaluation components can be swapped independently. Reproducibility means that different routing methods are executed under the same data loading, feature extraction, configuration, and logging protocol.

A typical experiment is specified by two types of configuration files. The benchmark configuration defines dataset-level settings, including the data source, modality, model pool, encoder, split protocol, budget setting, and random seed. The router configuration defines method-level settings, including training hyperparameters and algorithm-specific options. With this separation, changing the router does not alter the dataset pipeline, and changing the encoder does not require rewriting router logic.

A standard run follows a unified entry point:
\begin{verbatim}
python main.py --dataset [Dataset Name] --method [Method Name]
\end{verbatim}

The dataset name determines the data source, model pool, encoder, and system assumptions, while the method name determines the routing algorithm and its configuration. This allows researchers to rapidly test new routing methods while keeping all non-algorithmic factors controlled.

\subsection{Adding New Benchmarks or Datasets}

\textsc{ORBIT} allows users to add new benchmarks or private datasets while preserving the same routing pipeline. A new dataset only needs to follow the ORBIT data format and implement the required loading interface.

\begin{enumerate}[leftmargin=10pt]
    \item \textbf{Prepare the dataset in ORBIT format.}
    The dataset should define the query set, candidate model pool, available model outputs or performance records, and the corresponding cost information when applicable. For RouteJudge-oriented development, the dataset may also contain historical preference records or model comparison outcomes.

    \item \textbf{Register the dataset loader.}
    Add the dataset loader to \texttt{utils/data.py} by implementing the required dataset base class. Once registered, the dataset can be loaded by the shared ORBIT pipeline and paired with any compatible encoder and router.

    \item \textbf{Define the benchmark configuration.}
    Create a configuration file under \texttt{configs/benchmarks/} to specify the modality, model pool, split setting, encoder choice, budget setting, and other dataset-level options. If a router requires model descriptions or auxiliary metadata, the corresponding files can be placed under \texttt{configs/description/}.
\end{enumerate}

After registration, the new dataset can be used with existing routers without modifying router implementations. This makes it possible to test whether a routing method generalizes across different query distributions, model pools, and deployment assumptions.

\subsection{Adding New Encoders}

\textsc{ORBIT} treats query representation as a replaceable component. New encoders can be added to the shared embedding layer and reused by all compatible routers.

To add a new encoder, users register it in \texttt{utils/embedding.py}. Both unimodal and multimodal encoders are supported. The encoder can then be selected in the benchmark configuration by setting the \texttt{embeddings} field. Once registered, the same encoder can be used across different routers, enabling controlled studies of how query representation affects routing behavior.

This design is useful for two types of research. First, researchers can test whether stronger or domain-specific embeddings improve routing decisions. Second, they can study the trade-off between encoder overhead and routing quality, which is important when routing itself must remain lightweight.

\subsection{Adding New Routing Algorithms}

The main extension point of \textsc{ORBIT} is the router interface. A new routing method can be integrated with minimal boilerplate while reusing the same data pipeline, embedding layer, training utilities, and evaluation scripts.

\begin{enumerate}[leftmargin=10pt]
    \item \textbf{Implement the router logic.}
    Create a new router class under \texttt{methods/} by inheriting from \texttt{BaseRouter}. The base class provides shared functionality for data loading, embedding extraction, training, checkpointing, and prediction. The new method only needs to implement its training and inference routines.

    \item \textbf{Expose a unified prediction interface.}
    The router should take the query representation and the feasible model set as input, and output the selected model. This interface is required for both offline ORBIT evaluation and online RouteJudge deployment.

    \item \textbf{Register and configure the router.}
    Register the method in \texttt{train.py} and add a configuration file under \texttt{configs/routers/}. The configuration should include method-specific hyperparameters, training options, and any required auxiliary files.

    \item \textbf{Check reproducibility.}
    The implementation should be runnable through the standard ORBIT command-line entry point. All required dependencies, random seeds, checkpoints, and preprocessing steps should be specified clearly so that the method can be reproduced by the maintainers.
\end{enumerate}

Once these steps are completed, the new router can be evaluated within ORBIT under the same protocol as existing methods. More importantly, it becomes eligible for RouteJudge integration.

\subsection{Evaluating New Routers on RouteJudge through ORBIT}

\textsc{ORBIT} also serves as the submission and integration layer for RouteJudge. Researchers who want their routing methods to be evaluated on RouteJudge can submit their ORBIT-compatible implementations through a pull request. The submitted code should be directly runnable, include the required router configuration, and follow the unified prediction interface.

The RouteJudge integration process consists of two stages.

\paragraph{Stage 1: Historical replay evaluation.}
After receiving a compatible router submission, we first train or initialize the router within ORBIT using the available training protocol. We then evaluate it on historical RouteJudge records. Each historical record contains the user query, the compared model pair, and the user preference label. If the submitted router selects the model that was preferred by the user in the historical comparison, the router receives a historical win. If the historical label is \texttt{Tie} or \texttt{Both Bad}, the attribution follows the same scoring rule used by RouteJudge. This replay stage provides an initial estimate of whether the submitted router is aligned with past user preferences before it enters live online evaluation.

\paragraph{Stage 2: Online RouteJudge evaluation.}
After historical replay, the router is added to the RouteJudge router pool for online testing. In live evaluation, it is treated in the same way as existing routers: for each user query, it recommends a model under the same model pool and budget constraints, participates in the same anonymous pairwise comparison protocol, and receives preference attribution when its selected model enters the judged duel pair. Its results are then updated on RouteJudge using the same metrics as other routers, including preference score, win rate, Elo rating, cost-aware performance, participation rate, and task-conditioned behavior.

Because each submitted router must be checked, trained, validated, and deployed, there may be a delay between pull request submission and RouteJudge evaluation. In the current workflow, this delay is typically between two and seven days, depending on implementation complexity, dependency issues, training cost, and the online evaluation queue. Once the router has been evaluated, researchers can query its results on the RouteJudge platform and compare it with existing routing methods.

This submission pipeline makes RouteJudge an open and continuously expandable routing evaluation platform. Researchers can develop methods in ORBIT, validate them offline, submit compatible routers through pull requests, and obtain both historical replay results and online user-preference results on RouteJudge.
\section{Preliminary Results}
\label{sec:results}

We present preliminary results from both the offline \textsc{ORBIT} pipeline and the online \textbf{RouteJudge} platform. The goal of these experiments is not to provide a final ranking of all routing methods, but to illustrate the two complementary roles of our ecosystem: \textsc{ORBIT} supports reproducible offline development and comparison, while \textbf{RouteJudge} evaluates router decisions under real user preferences.

\subsection{Offline Benchmark Evaluation with ORBIT}
\label{sec:orbit_results}

\begin{figure*}[t]
  \centering
  \begin{subfigure}[t]{0.48\linewidth}
    \centering
    \includegraphics[width=\linewidth]{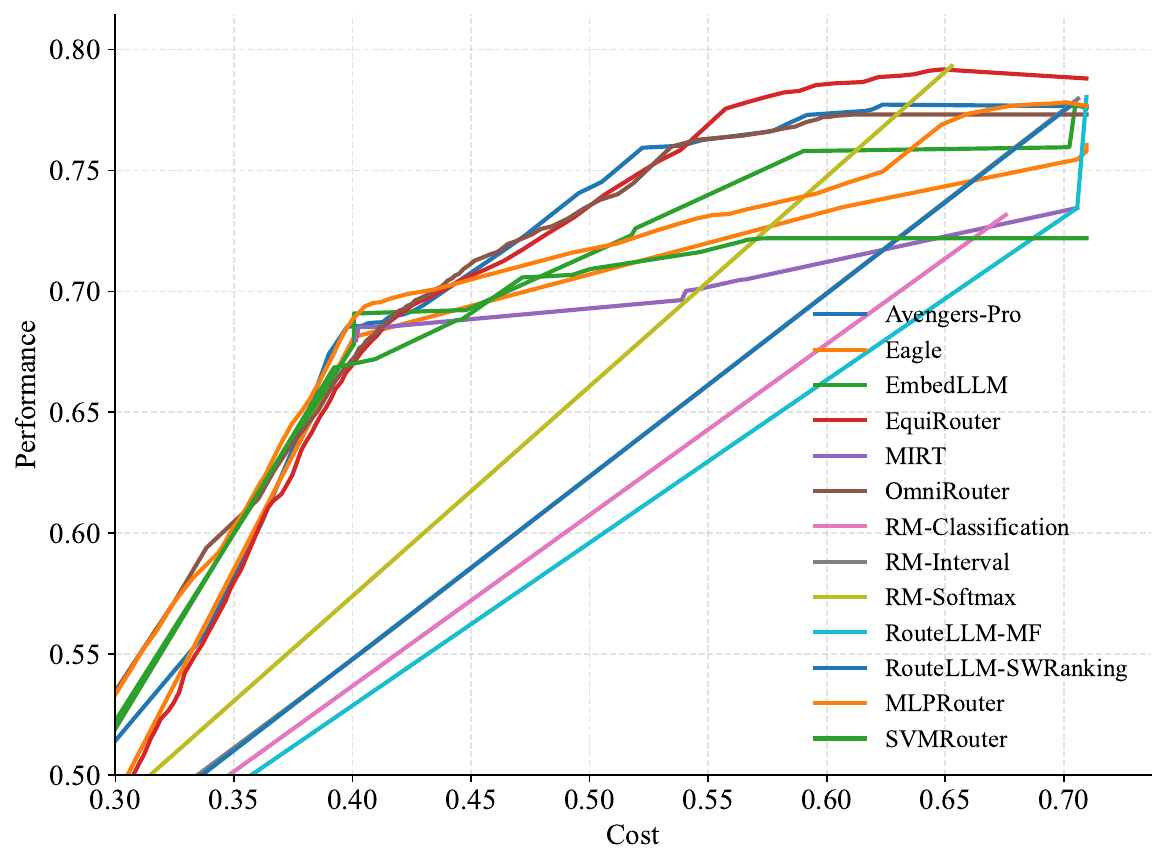}
    \caption{
    Performance--cost trade-off curves of representative routers on RouterEval under the unified \textsc{ORBIT} pipeline.
    }
    \label{fig:routereval}
  \end{subfigure}
  \hfill
  \begin{subfigure}[t]{0.48\linewidth}
    \centering
    \includegraphics[width=\linewidth]{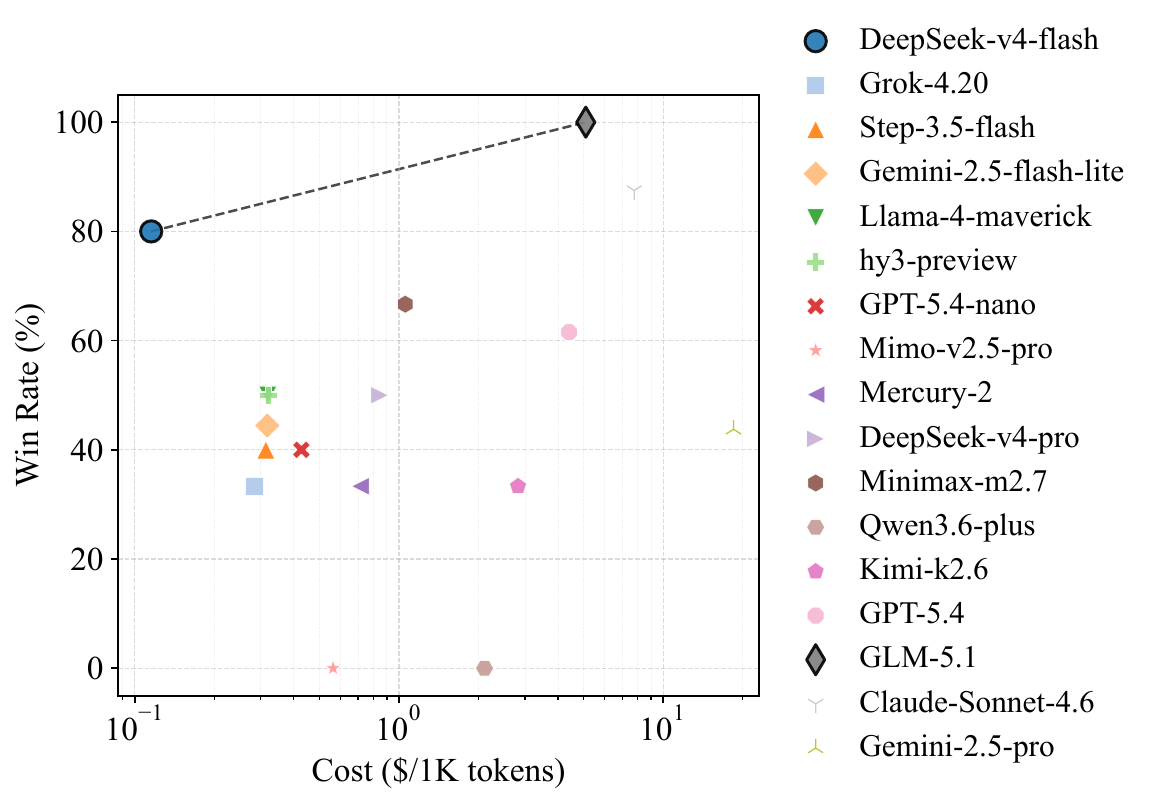}
    \caption{
    Model-level cost--win-rate distribution on \textbf{RouteJudge}.
    }
    \label{fig:model_cost_winrate}
  \end{subfigure}
  \caption{
  Preliminary results from the offline \textsc{ORBIT} pipeline and the online \textbf{RouteJudge} platform.
  }
  \label{fig:preliminary_results}
\end{figure*}

We first evaluate representative routing methods using the unified \textsc{ORBIT} pipeline. We use the RouterEval benchmark and adopt a fixed experimental configuration across all routing methods. Queries are encoded using \texttt{all-MiniLM-L6-v2}, and we use a fixed train-test split ratio of 2:8 to reflect a low-data setting where routers are trained with limited supervision. We then sweep the budget to obtain the performance--cost trade-off curves shown in Figure~\ref{fig:routereval}.

These results demonstrate the main purpose of \textsc{ORBIT}: different routers can be evaluated under the same data processing, query representation, model pool, budget setting, and reporting protocol. This removes common sources of inconsistency across implementations and ensures that observed differences mainly reflect routing strategies rather than mismatched preprocessing, encoders, or evaluation scripts. In addition to the trade-off curves, \textsc{ORBIT} also logs curve-level metrics such as nAUC, Peak Score, QNC, and RCI, which are provided in the appendix.

The same pipeline also supports rapid component-level ablation. Since the benchmark configuration is shared across methods, changing the encoder, data split, budget grid, or model pool only requires modifying the corresponding configuration entry, and the change is automatically applied to all compatible routers. This composability is important for routing research, where conclusions can otherwise be sensitive to implementation-specific choices.

\subsection{Online Preference Evaluation with RouteJudge}
\label{sec:routejudge_results}

We next report preliminary results collected from the \textbf{RouteJudge} platform as of \textbf{2026-06-08 04:00 AoE} (UTC-12). At that time, the platform had recorded 261 matches and 109 user votes. The following analysis is based on these 109 voted comparisons. Since the number of votes is still limited, the results should be interpreted as an initial empirical snapshot rather than a definitive ranking of routing strategies. Continuously updated platform statistics are available at \url{https://routejudge.cn/stats}.

\subsubsection{Router Ranking}

Table~\ref{tab:router_elo_results} reports the current router ranking by Elo score. Overall, the leading routers obtain both higher Elo ratings and above-random preference win rates, suggesting that online user preferences can already differentiate routing strategies at this early stage. RouterLLM-MF obtains the highest Elo score, while NIRT-Router achieves the highest observed win rate. Several embedding-based, graph-based, retrieval-based, and regression-style routers also remain competitive, indicating that different routing paradigms can be effective under user-preference-based evaluation.

\definecolor{wincolor}{RGB}{0,120,70}
\definecolor{losecolor}{RGB}{180,50,45}
\definecolor{neutralcolor}{RGB}{100,100,100}

\newcommand{\winratepos}[1]{\textcolor{wincolor}{#1}}
\newcommand{\winrateneg}[1]{\textcolor{losecolor}{#1}}
\newcommand{\winrateneu}[1]{\textcolor{neutralcolor}{#1}}

\begin{table}[t]
\centering
\caption{
Router performance ranked by Elo score. Results are based on 109 user-voted comparisons collected by \textbf{RouteJudge}.
}
\label{tab:router_elo_results}
\footnotesize
\setlength{\tabcolsep}{3pt}
\renewcommand{\arraystretch}{0.92}
\begin{tabularx}{\columnwidth}{@{}>{\raggedright\arraybackslash}X
                                >{\centering\arraybackslash}p{0.20\columnwidth}
                                >{\centering\arraybackslash}p{0.16\columnwidth}@{}}
\toprule
Router & Win Rate & Elo \\
\midrule
RouterLLM-MF \cite{ong2024routellm} & \winratepos{66.67\%} & \textbf{1278} \\
NIRT-Router \cite{song2025irt} & \winratepos{\textbf{80.00\%}} & 1274 \\
kNNRouter \cite{stripelis2024tensoropera} & \winratepos{60.00\%} & 1240 \\
GraphRouter \cite{feng2025graphrouter} & \winrateneg{44.44\%} & 1218 \\
EmbedLLM \cite{zhuang2024embedllm} & \winratepos{63.64\%} & 1215 \\
EquiRouter \cite{lai2026routing} & \winratepos{64.00\%} & 1212 \\
RouterDC \cite{chen2024routerdc} & \winratepos{57.14\%} & 1209 \\
SVMRouter \cite{stripelis2024tensoropera} & \winratepos{51.72\%} & 1205 \\
Avengers-Pro \cite{zhang2025beyond} & \winrateneu{50.00\%} & 1204 \\
Avengers \cite{zhang2025avengers} & \winratepos{58.33\%} & 1198 \\
RM-Interval \cite{tsiourvas2025causal} & \winrateneg{46.15\%} & 1184 \\
MLPRouter \cite{stripelis2024tensoropera} & \winrateneg{39.13\%} & 1176 \\
Eagle \cite{zhao2024eagle} & \winrateneg{43.48\%} & 1175 \\
OmniRouter \cite{mei2025omnirouter} & \winrateneg{40.00\%} & 1170 \\
RM-CLS \cite{tsiourvas2025causal} & \winrateneg{46.43\%} & 1169 \\
RouteLLM-SW \cite{ong2024routellm} & \winrateneg{27.27\%} & 1145 \\
RM-Softmax \cite{tsiourvas2025causal} & \winrateneg{31.82\%} & 1136 \\
HybridLLM \cite{ding2024hybrid} & \winrateneg{18.18\%} & 1134 \\
MIRT-Router \cite{song2025irt} & \winrateneg{36.00\%} & 1117 \\
\bottomrule
\end{tabularx}
\end{table}

The ranking should be interpreted together with the number of available comparisons and the participation behavior of each router. A router with a high win rate may have participated in fewer decisive comparisons, while a router with a lower win rate may have been evaluated under more diverse or more difficult query conditions. This is why \textbf{RouteJudge} reports Elo, win rate, participation, and cost-related statistics jointly rather than relying on a single aggregate score.

The preliminary results also suggest that strong offline routing designs do not necessarily transfer uniformly to online preference-based evaluation. Some routers with explicit learned scoring mechanisms obtain relatively low preference win rates, while simpler non-parametric or matrix-factorization-based methods remain competitive. This observation supports the motivation of \textbf{RouteJudge}: router quality should not be assessed only by agreement with offline labels or static benchmark scores, but also by whether the selected models produce responses preferred by users under real interaction settings.

\subsubsection{Model Cost--Preference Trade-off}
\label{sec:model_pareto}

We further examine the model-level cost--preference trade-off in \textbf{RouteJudge}. Figure~\ref{fig:model_cost_winrate} plots each candidate model by its average inference cost and empirical preference win rate. The dashed line denotes the observed Pareto frontier, where no model is simultaneously cheaper and preferred more often.

The results show that model preference is not determined by cost alone. Some low-cost models achieve competitive win rates, while several stronger models require substantially higher budgets. This suggests that effective routing should adapt model selection to the user-selected budget and task context, rather than always choosing the most expensive or highest-capability model.

\subsection{Discussion}

Taken together, the offline and online results highlight the complementary roles of \textsc{ORBIT} and \textbf{RouteJudge}. \textsc{ORBIT} provides a controlled environment for developing and comparing routers under reproducible benchmark settings, while \textbf{RouteJudge} evaluates whether those routing decisions align with real user preferences. The gap between offline benchmark performance and online preference outcomes further motivates our two-stage evaluation design: routers should first be validated under standardized offline protocols, and then tested under user-facing preference feedback.

\section{Conclusion}
\label{sec:conclusion}

We presented \textbf{RouteJudge}, an online pairwise preference evaluation platform for LLM routing systems, together with \textsc{ORBIT}, a modular toolbox for router development, offline benchmarking, and submission-based integration. \textbf{RouteJudge} evaluates router-level decision quality by attributing anonymous user preference feedback to the routing strategies behind compared model responses. \textsc{ORBIT} provides the standardized development layer needed to implement, validate, and submit new routing algorithms under consistent protocols.

Together, \textbf{RouteJudge} and \textsc{ORBIT} form an open evaluation ecosystem for LLM routing: researchers can develop routers in \textsc{ORBIT}, compare them on existing benchmarks, submit compatible methods for historical replay and online testing, and track their performance on \textbf{RouteJudge}. Preliminary results show that offline reproducible evaluation and online preference-based evaluation provide complementary signals, suggesting that future routing research should consider not only benchmark-optimal model selection, but also user-preferred, cost-aware, and deployment-sensitive routing behavior.

\vskip 0.2in
\bibliography{sample}

\clearpage
\appendix

\setcounter{figure}{0}
\setcounter{table}{0}
\setcounter{equation}{0}
\renewcommand{\thefigure}{\Alph{section}.\arabic{figure}}
\renewcommand{\thetable}{\Alph{section}.\arabic{table}}
\renewcommand{\theequation}{\Alph{section}.\arabic{equation}}

\section{Candidate Models and Routing Strategies}
\label{app:routers_and_models}

This appendix provides additional details about the candidate model pool and routing strategies used in the current RouteJudge platform and supported by the ORBIT toolbox. RouteJudge evaluates routers over a shared model pool under the same budget-feasible model space, ensuring that performance differences are attributed to routing decisions rather than access to different candidate models.

\subsection{Candidate Model Pool}

The current RouteJudge platform includes 17 candidate models. The pool contains both low-cost models for budget-sensitive routing and stronger models for quality-sensitive queries:

\begin{itemize}[leftmargin=*]
    \item \textbf{DeepSeek-v4-flash}: A low-cost candidate model used for budget-sensitive routing decisions.
    \item \textbf{Grok-4.20}: A general-purpose candidate model included in the shared model pool.
    \item \textbf{Step-3.5-flash}: A lightweight candidate model used for efficient response generation.
    \item \textbf{Gemini-2.5-flash-lite}: A low-cost Gemini-family model included for cost-efficient routing.
    \item \textbf{Llama-4-maverick}: An open-family candidate model used as part of the heterogeneous model pool.
    \item \textbf{hy3-preview}: A preview candidate model included to increase model diversity.
    \item \textbf{GPT-5.4-nano}: A compact GPT-family model used for low-cost query handling.
    \item \textbf{Mimo-v2.5-pro}: A pro-level candidate model included for stronger response generation.
    \item \textbf{Mercury-2}: A general candidate model included in the RouteJudge model pool.
    \item \textbf{DeepSeek-v4-pro}: A stronger DeepSeek-family model used for higher-quality routing choices.
    \item \textbf{Minimax-m2.7}: A general-purpose candidate model included for comparison across model families.
    \item \textbf{Qwen3.6-plus}: A Qwen-family model used as a higher-capability candidate in the model pool.
    \item \textbf{Kimi-k2.6}: A Kimi-family candidate model included for diverse generation behavior.
    \item \textbf{GPT-5.4}: A high-capability GPT-family model used for quality-oriented routing decisions.
    \item \textbf{GLM-5.1}: A GLM-family candidate model included in the shared model pool.
    \item \textbf{Claude-Sonnet-4.6}: A high-capability Claude-family model used for quality-sensitive queries.
    \item \textbf{Gemini-2.5-pro}: A pro-level Gemini-family model included for high-quality response generation.
\end{itemize}

\subsection{Routing Strategies in RouteJudge}

RouteJudge includes routing strategies from multiple families, including rule-based, regression-based, classification-based, ranking-based, graph-based, reward-model-based, cascade-based, and non-parametric methods. The current implementation contains the following routers:

\begin{itemize}[leftmargin=*]
    \item \textbf{RouterLLM-MF}~\citep{ong2024routellm}: A matrix-factorization-based RouteLLM variant that estimates model suitability from observed routing data.
    \item \textbf{RouteLLM-SW}~\citep{ong2024routellm}: A RouteLLM variant that uses similarity- or score-weighted information for model selection.
    \item \textbf{NIRT-Router}~\citep{song2025irt}: An item-response-theory-based router that models query difficulty and model ability.
    \item \textbf{MIRT-Router}~\citep{song2025irt}: A multidimensional IRT-based router that represents model ability and query difficulty with multiple latent factors.
    \item \textbf{kNNRouter}~\citep{stripelis2024tensoropera}: A non-parametric router that selects models according to nearest historical examples.
    \item \textbf{SVMRouter}~\citep{stripelis2024tensoropera}: A classification-based router that predicts a suitable model from query features.
    \item \textbf{MLPRouter}~\citep{stripelis2024tensoropera}: A neural classification-based router that maps query representations to model choices.
    \item \textbf{GraphRouter}~\citep{feng2025graphrouter}: A graph-based router that exploits structured relations among models, tasks, or queries.
    \item \textbf{EmbedLLM}~\citep{zhuang2024embedllm}: An embedding-based router that estimates model suitability by comparing queries in representation space.
    \item \textbf{EquiRouter}~\citep{lai2026routing}: A decision-aware routing strategy that estimates relative model utility under routing constraints.
    \item \textbf{RouterDC}~\citep{chen2024routerdc}: A data-driven router that learns model-selection decisions from offline routing signals.
    \item \textbf{Avengers}~\citep{zhang2025avengers}: An ensemble-style router that combines multiple decision signals for model selection.
    \item \textbf{Avengers-Pro}~\citep{zhang2025beyond}: An enhanced Avengers-style router with stronger aggregation or decision mechanisms.
    \item \textbf{Eagle}~\citep{zhao2024eagle}: A cascade-style router that decides whether to invoke stronger models based on estimated difficulty or uncertainty.
    \item \textbf{OmniRouter}~\citep{mei2025omnirouter}: A general-purpose router designed for model selection across heterogeneous task types.
    \item \textbf{HybridLLM}~\citep{ding2024hybrid}: A hybrid cost--quality-aware router that combines multiple routing criteria.
    \item \textbf{RM-CLS}~\citep{tsiourvas2025causal}: A reward-model-based classification router that predicts the preferred model.
    \item \textbf{RM-Softmax}~\citep{tsiourvas2025causal}: A reward-model-based router that selects models according to softmax-normalized preference scores.
    \item \textbf{RM-Interval}~\citep{tsiourvas2025causal}: A reward-model-based router that incorporates interval-style uncertainty in routing decisions.
\end{itemize}

\subsection{Additional Routers Supported by ORBIT}

In addition to the routers currently active in RouteJudge, ORBIT supports additional routing methods that can be used for offline evaluation or future online deployment:

\begin{itemize}[leftmargin=*]
    \item \textbf{RouteLLM-BERT}~\citep{ong2024routellm}: A BERT-based classifier trained on pairwise preference data to predict which model will be preferred for a given prompt.
    \item \textbf{ModelSAT}~\citep{zhang2025capability}: A capability-aware router that uses structured model capability representations for dynamic model selection.
    \item \textbf{Oracle}: A non-deployable upper bound that selects the best feasible model using ground-truth outcomes.
\end{itemize}

\section{ORBIT Details}
\label{app:orbit_details}

This appendix provides the technical details of ORBIT that are omitted from the main text, including the formal routing setup, benchmark suite, embedding interfaces, and offline evaluation protocol.

\subsection{Budgeted Routing Formulation}

Let the candidate model pool be $\mathcal{M}=\{1,\dots,K\}$, where each index $j\in\mathcal{M}$ denotes a deployed LLM with its own capability, cost, latency, context length, modality support, or serving characteristics. Queries are sampled from a benchmark or application distribution over a query space $\mathcal{Q}$, denoted by $q\sim\mathcal{D}$.

For each query $q$ and model $j$, we define two query-dependent quantities: a performance or utility value $a_j(q)\in\mathbb{R}$, measuring the quality of model $j$ on $q$, and an inference cost $c_j(q)\in\mathbb{R}_{+}$, measuring resources consumed by running model $j$ on $q$. The cost may correspond to monetary cost, latency, or a composite resource measure.

A router usually cannot access $a_j(q)$ at decision time. Instead, it derives a query representation through an embedding or feature function
\[
\phi:\mathcal{Q}\rightarrow\mathbb{R}^{d},
\qquad
x=\phi(q).
\]
Given a budget $C$, the budget-feasible model set is
\begin{equation}
\mathcal{F}(q;C)=\{\,j\in\mathcal{M}\mid c_j(q)\le C\,\}.
\end{equation}
A budgeted routing policy is a function $\pi_C:\mathbb{R}^{d}\rightarrow\mathcal{M}$ that maps the query representation to a selected model. The offline budgeted routing objective is
\begin{equation}
\begin{aligned}
\max_{\pi_C}\quad &
\mathbb{E}_{q\sim\mathcal{D}}
\left[
a_{\pi_C(\phi(q))}(q)
\right] \\
\text{s.t.}\quad &
\pi_C(\phi(q))\in \mathcal{F}(q;C),
\quad \forall q .
\end{aligned}
\label{eq:app_routing_problem}
\end{equation}
This formulation captures the main offline routing problem studied in ORBIT: selecting a budget-feasible model that maximizes expected performance under a fixed evaluation protocol.

\subsection{Benchmark Suite and Splits}

ORBIT integrates both unimodal and multimodal routing benchmarks. For unimodal routing, ORBIT supports RouterBench~\citep{hu2024routerbench}, MixInstruct~\citep{llm-blender-2023}, and RouterEval~\citep{huang2025routereval}. For multimodal routing, ORBIT supports MMR-Bench~\citep{ma2026mmrbench}. These benchmarks provide query distributions, candidate model pools, model outputs or performance records, and cost information that can be used to train and evaluate routing methods under controlled settings.

To evaluate both average-case performance and robustness, ORBIT supports in-domain and out-of-domain splits. The in-domain split follows the same distribution as router development data, while the out-of-domain split holds out domains, tasks, or modality patterns that differ from the development distribution. All benchmarks are loaded through the same ORBIT data interface so that routing methods can be compared under consistent preprocessing and reporting protocols.

\subsection{Embedding Layer}

ORBIT provides a unified embedding layer that maps each query into a feature representation used by routers. The embedding function $\phi(\cdot)$ can be instantiated by lightweight text encoders, larger embedding models, or multimodal encoders for image-text inputs. This design separates query representation from router implementation, allowing researchers to swap encoders without modifying router logic.

ORBIT also supports end-to-end methods that jointly train the encoder and routing policy. In this case, the embedding module can be optimized together with the router to better capture query difficulty, task type, and model-specific compatibility.

\subsection{Offline Evaluation Protocol}

ORBIT evaluates routing methods by sweeping the per-query budget and tracing a performance--cost trade-off curve. For a routing policy under budget $C$, denoted by $\pi_C$, the expected test performance and expected inference cost are
\[
A(C)=
\mathbb{E}_{q\sim\mathcal{D}_{\mathrm{test}}}
\left[
a_{\pi_C(q)}(q)
\right],
\qquad
\bar{c}(C)=
\mathbb{E}_{q\sim\mathcal{D}_{\mathrm{test}}}
\left[
c_{\pi_C(q)}(q)
\right].
\]
Evaluating these quantities over a grid of budgets gives the trade-off curve
\[
\left\{(\bar{c}(C),A(C))\right\}.
\]
ORBIT reports curve-level metrics to summarize different aspects of routing behavior.

\paragraph{Normalized Area Under the Curve.}
Let budgets be ordered by increasing cost, and define $x_t=\bar{c}(C_t)$ and $y_t=A(C_t)$. The normalized area under the performance--cost curve is
\begin{equation}
\mathrm{nAUC}
=
\frac{1}{x_T-x_1}
\sum_{t=1}^{T-1}
\frac{y_t+y_{t+1}}{2}
(x_{t+1}-x_t).
\end{equation}

\paragraph{Peak Score.}
The peak performance on the curve is
\begin{equation}
P_s
=
\max_{t\in\{1,\dots,T\}} y_t,
\qquad
x_{t^\star}
\ \text{with}\
t^\star\in\arg\max_t y_t .
\end{equation}

\paragraph{Quality Neutral Cost.}
Let
\[
A_{\max}
=
\max_{j\in\mathcal{M}}
\mathbb{E}_{q\sim\mathcal{D}_{\mathrm{test}}}
[a_j(q)]
\]
denote the performance of the best single model, and let $x_{\max}$ be its corresponding average cost. The quality neutral cost is defined as
\begin{equation}
\mathrm{QNC}
=
\min_{t:\,y_t\ge A_{\max}} x_t,
\qquad
\mathrm{rQNC}
=
\mathrm{QNC}/x_{\max}.
\end{equation}

\paragraph{Routing Collapse Index.}
For each query $q$, define the best achievable performance and the cheapest cost among best models as
\[
a^\star(q)=\max_{j\in\mathcal{M}}a_j(q),
\qquad
c^\star(q)=
\min_{j:\,a_j(q)=a^\star(q)}c_j(q).
\]
The routing collapse indicator is
\begin{equation}
\mathrm{RCI}(q)
=
\mathbbm{1}
\left[
a_{\pi_C(q)}(q)<a^\star(q)
\ \text{or}\
c_{\pi_C(q)}(q)>c^\star(q)
\right],
\qquad
\mathrm{RCI}
=
\mathbb{E}_{q\sim\mathcal{D}_{\mathrm{test}}}
[\mathrm{RCI}(q)].
\end{equation}

\section{RouteJudge Evaluation Metrics}
\label{app:routejudge_metrics}

This appendix details the metrics used by RouteJudge to evaluate routing strategies from preference, cost, task-conditioned, and diagnostic perspectives.

Let $\mathcal{D}_{\mathrm{RJ}}=\{\mathcal{Z}_t\}_{t=1}^{T}$ denote the set of RouteJudge evaluation records. For the $t$-th record, $m_i^{(t)}$ denotes the model selected by router $r_i$, $(m_A^{(t)},m_B^{(t)})$ denotes the selected duel pair, $y^{(t)}$ denotes the user preference label, $S_i^{(t)}\in\{0,0.5,1,\varnothing\}$ denotes the attributed score of router $r_i$, and $\tau^{(t)}$ denotes the task label.

\subsection{Preference-Based Metrics}

Since a router may recommend a model that does not enter the final pairwise comparison, we define the participating set of router $r_i$ as
\begin{equation}
\mathcal{P}_i
=
\{t\mid S_i^{(t)}\ne\varnothing\}.
\end{equation}
The participation rate is
\begin{equation}
\mathrm{PartRate}(r_i)
=
\frac{|\mathcal{P}_i|}{T}.
\end{equation}

The average preference score is
\begin{equation}
\mathrm{PrefScore}(r_i)
=
\frac{1}{|\mathcal{P}_i|}
\sum_{t\in\mathcal{P}_i}
S_i^{(t)}.
\end{equation}
Wins receive score $1$, ties receive $0.5$, losses receive $0$, and \texttt{Both Bad} assigns $0$ to both participating routers.

For comparisons with a clear winner, we define
\[
\mathcal{B}_i
=
\{t\in\mathcal{P}_i
\mid
y^{(t)}\in\{\text{\texttt{A Win}},\text{\texttt{B Win}}\}
\}.
\]
The binary win rate is
\begin{equation}
\mathrm{WinRate}(r_i)
=
\frac{
\sum_{t\in\mathcal{B}_i}
\mathbbm{1}[S_i^{(t)}=1]
}{
|\mathcal{B}_i|
}.
\end{equation}
RouteJudge also reports Elo ratings and bootstrap confidence intervals when enough comparisons are available.

\subsection{Cost--Preference Trade-off}

For router $r_i$, the average participating cost is
\begin{equation}
\mathrm{Cost}(r_i)
=
\frac{1}{|\mathcal{P}_i|}
\sum_{t\in\mathcal{P}_i}
c^{(t)}(m_i^{(t)}),
\end{equation}
where $c^{(t)}(m_i^{(t)})$ denotes the actual inference cost of the model selected by router $r_i$ in the $t$-th comparison.

RouteJudge analyzes routers in the cost--preference space, where the quality axis can be $\mathrm{PrefScore}(r_i)$ or $\mathrm{WinRate}(r_i)$ and the cost axis is $\mathrm{Cost}(r_i)$. A router $r_i$ dominates another router $r_j$ if
\[
\mathrm{PrefScore}(r_i)\ge \mathrm{PrefScore}(r_j),
\qquad
\mathrm{Cost}(r_i)\le \mathrm{Cost}(r_j),
\]
with at least one inequality being strict. Routers on the Pareto frontier represent strategies that achieve the best observed preference performance at a given cost level.

\subsection{Task-Conditioned Metrics}

For a task category $\tau$, define the task-conditioned participating set as
\begin{equation}
\mathcal{P}_{i,\tau}
=
\{t\in\mathcal{P}_i\mid \tau^{(t)}=\tau\}.
\end{equation}
The task-conditioned preference score is
\begin{equation}
\mathrm{PrefScore}(r_i\mid\tau)
=
\frac{1}{|\mathcal{P}_{i,\tau}|}
\sum_{t\in\mathcal{P}_{i,\tau}}
S_i^{(t)}.
\end{equation}
The task-conditioned win rate is computed analogously by restricting $\mathcal{B}_i$ to examples with task label $\tau$.

\subsection{Pairwise Router Comparison}

For two routers $r_i$ and $r_j$, define their shared participating set as
\begin{equation}
\mathcal{P}_{ij}
=
\mathcal{P}_i\cap\mathcal{P}_j.
\end{equation}
The head-to-head preference score of $r_i$ against $r_j$ is
\begin{equation}
\mathrm{H2H}(r_i,r_j)
=
\frac{1}{|\mathcal{P}_{ij}|}
\sum_{t\in\mathcal{P}_{ij}}
\varphi(S_i^{(t)},S_j^{(t)}),
\end{equation}
where
\[
\varphi(a,b)
=
\begin{cases}
1, & a>b,\\
0.5, & a=b,\\
0, & a<b.
\end{cases}
\]

When binary preference outcomes are available, RouteJudge can also apply McNemar's test to compare two routers on shared examples. Let $a$ be the number of shared comparisons where $r_i$ wins and $r_j$ loses, and let $b$ be the number of shared comparisons where $r_j$ wins and $r_i$ loses. The test statistic is
\begin{equation}
\chi^2
=
\frac{(|a-b|-1)^2}{a+b}.
\end{equation}

\subsection{Routing Behavior Diagnostics}

RouteJudge further analyzes routing behavior beyond aggregate scores.

\paragraph{Routing entropy.}
Let $p_{i,m}$ denote the empirical frequency with which router $r_i$ selects model $m$:
\[
p_{i,m}
=
\frac{1}{T}
\sum_{t=1}^{T}
\mathbbm{1}[m_i^{(t)}=m].
\]
The selection entropy is
\begin{equation}
H(r_i)
=
-
\sum_{m\in\mathcal{M}}
p_{i,m}\log_2 p_{i,m},
\qquad
H_{\mathrm{norm}}(r_i)
=
\frac{H(r_i)}{\log_2|\mathcal{M}|}.
\end{equation}

\paragraph{Router agreement.}
For two routers $r_i$ and $r_j$, the raw agreement rate is
\begin{equation}
\mathrm{Agree}(r_i,r_j)
=
\frac{1}{T}
\sum_{t=1}^{T}
\mathbbm{1}[m_i^{(t)}=m_j^{(t)}].
\end{equation}
Cohen's Kappa is
\begin{equation}
\kappa(r_i,r_j)
=
\frac{P_o-P_e}{1-P_e},
\end{equation}
where $P_o=\mathrm{Agree}(r_i,r_j)$ and
\[
P_e
=
\sum_{m\in\mathcal{M}}
p_{i,m}p_{j,m}.
\]

\paragraph{Consensus strength.}
For the $t$-th record, the consensus strength is
\begin{equation}
\rho^{(t)}
=
\frac{
\max_{m\in\mathcal{M}_C^{(t)}} v^{(t)}(m)
}{N},
\end{equation}
where $N$ is the number of routers. This metric measures how strongly routers agree on the same model for a given query.

\subsection{Historical Replay for Submitted Routers}

For a newly submitted ORBIT-compatible router, RouteJudge first performs historical replay before online deployment. Given a historical RouteJudge record with compared models $(m_A,m_B)$ and preference-derived scores $(s_A,s_B)$, the submitted router selects a model $\hat{m}$. Its replay score is
\begin{equation}
S_{\mathrm{replay}}
=
\begin{cases}
s_A, & \hat{m}=m_A,\\
s_B, & \hat{m}=m_B,\\
\varnothing, & \hat{m}\notin\{m_A,m_B\}.
\end{cases}
\end{equation}
Thus, if the submitted router selects the model that was historically preferred by users, it receives a win. If it selects the other compared model, it receives the corresponding loss. If its selected model was not involved in the historical comparison, the record is treated as non-participating. After historical replay, the router can enter live online RouteJudge evaluation under the same protocol as existing routers.

\section{Additional RouteJudge Results}
\label{app:additional_results}

This appendix provides additional analyses of routing behavior in RouteJudge. We focus on two complementary aspects: how different routers distribute their selections over the candidate model pool, and how routers compare against each other under shared user-voted comparisons.

\subsection{Model Selection Distribution}

\begin{figure*}[htb]
  \centering
  \includegraphics[width=\linewidth]{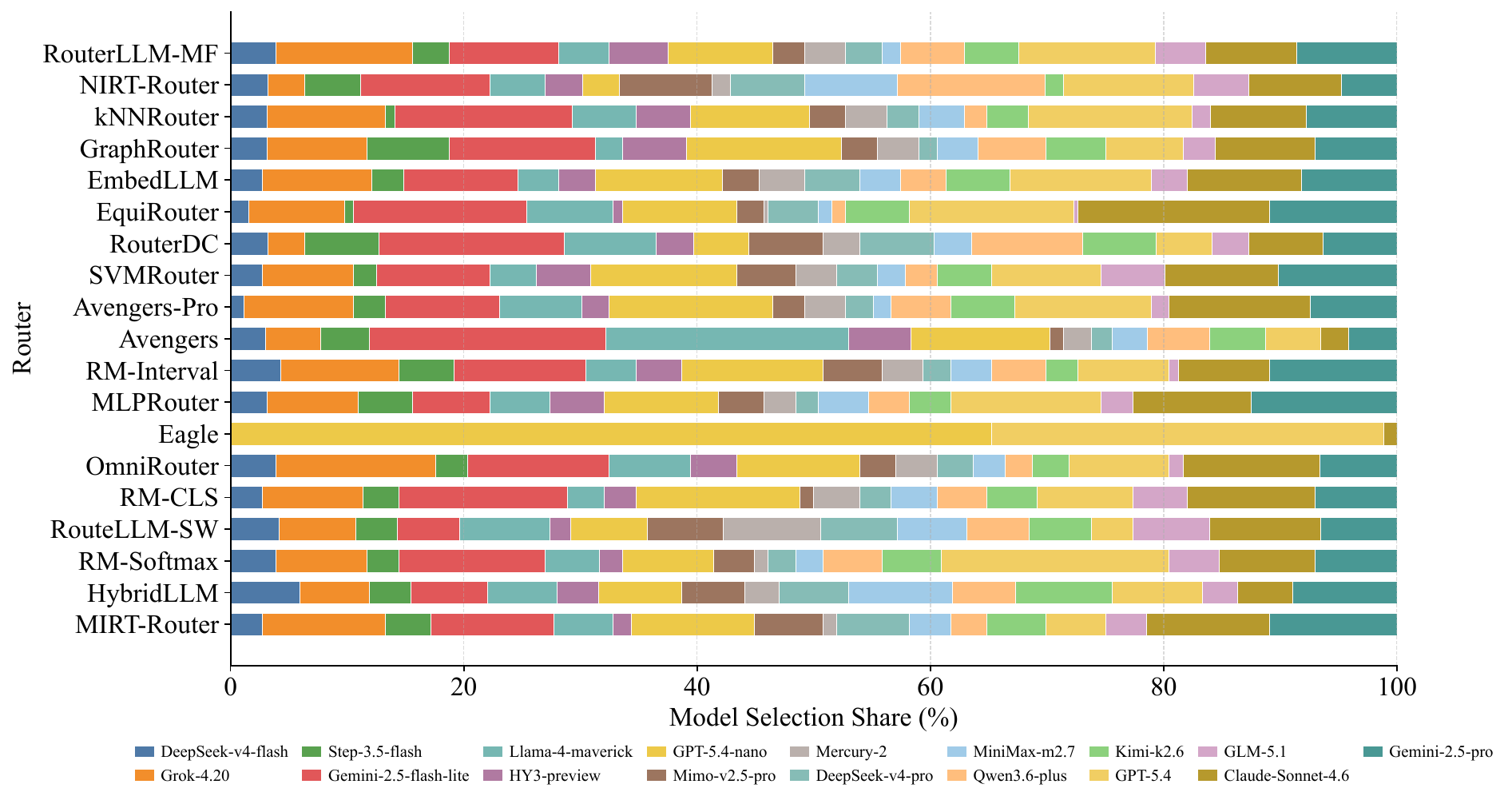}
  \caption{
  Model selection distribution of routing strategies. Each horizontal bar shows the percentage of times a router selects each candidate model. Concentrated distributions indicate conservative or model-specific routing behavior, while more diverse distributions suggest broader use of the candidate model pool.
  }
  \label{fig:router_model_selection_distribution}
\end{figure*}

Figure~\ref{fig:router_model_selection_distribution} shows the model selection distribution of each routing strategy. Each horizontal bar represents one router, and each segment indicates the fraction of queries for which the router selects a specific candidate model. The distribution reveals substantial differences in routing behavior. Some routers concentrate heavily on a small number of models, indicating more conservative or model-specific decision patterns. Other routers spread their selections across a broader range of models, suggesting stronger sensitivity to query-level differences or budget constraints.

This analysis complements the aggregate router ranking in the main text. A high-ranking router may obtain strong preference performance either by consistently selecting a robust model or by adapting its selections across different queries. Conversely, a router with diverse selections is not necessarily better unless this diversity leads to preferred responses. Therefore, model selection distribution provides a useful diagnostic for interpreting why routers obtain different win rates and Elo scores.

\subsection{Head-to-Head Router Comparison}

\begin{figure*}[htb]
  \centering
  \includegraphics[width=\linewidth]{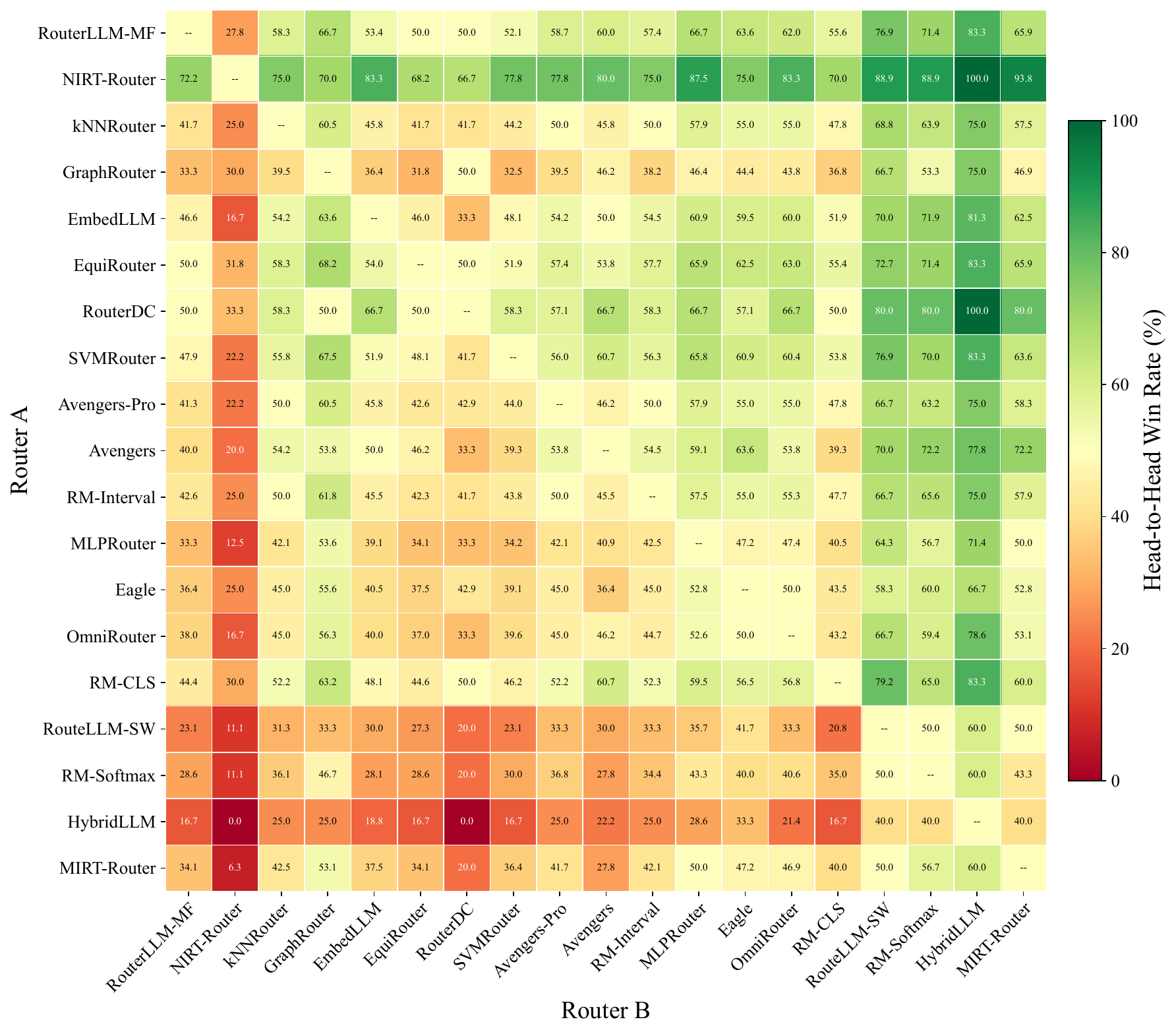}
  \caption{
  Head-to-head win-rate heatmap between routing strategies. Each cell reports the percentage of shared comparisons in which the row router outperforms the column router. Higher values indicate stronger pairwise advantage of the row router, while lower values indicate weaker relative performance against the corresponding column router.
  }
  \label{fig:router_head_to_head_heatmap}
\end{figure*}

Figure~\ref{fig:router_head_to_head_heatmap} presents the head-to-head win-rate matrix between routing strategies. Each cell reports the win rate of the row router against the column router on shared comparisons. Unlike global Elo ranking, this matrix directly shows pairwise advantages and disadvantages between routers. This is useful because two routers with similar overall scores may still behave differently on overlapping query subsets.

The heatmap shows that router performance is not uniformly ordered across all pairwise comparisons. Some routers exhibit broad advantages against many alternatives, while others perform competitively only against specific groups of routers. This suggests that RouteJudge can support a more fine-grained analysis than a single leaderboard score: it can reveal whether a router is consistently strong, specialized, or mainly competitive under certain comparison settings.

\end{document}